\documentclass[letterpaper]{article} 
\usepackage{aaai25}  
\usepackage{times}  
\usepackage{helvet}  
\usepackage{courier}  
\usepackage[hyphens]{url}  
\usepackage{graphicx} 
\usepackage{amsmath}
\usepackage{multirow}
\usepackage{amssymb}
\usepackage{natbib}  
\usepackage{caption} 
\usepackage{algorithm}
\usepackage{algorithmic}

\usepackage{newfloat}
\usepackage{listings}
\DeclareCaptionStyle{ruled}{labelfont=normalfont,labelsep=colon,strut=off} 
\floatstyle{ruled}
\newfloat{listing}{tb}{lst}{}
\floatname{listing}{Listing}
\nocopyright 

\title{More Pictures Say More: Visual Intersection Network for Open Set Object Detection}
\author {
    Bingcheng Dong\thanks{Equal contribution.}
    ~ Yuning Ding\footnotemark[1],
    ~ Jinrong Zhang,
    ~ Sifan Zhang,
    ~ Shenglan Liu\thanks{Corresponding author.}\\
    \textnormal{\normalsize Dalian University of Technology}
}

\usepackage{bibentry}

\begin{document}
\maketitle
\begin{figure*}[t]
\centering
\includegraphics[width=\textwidth]{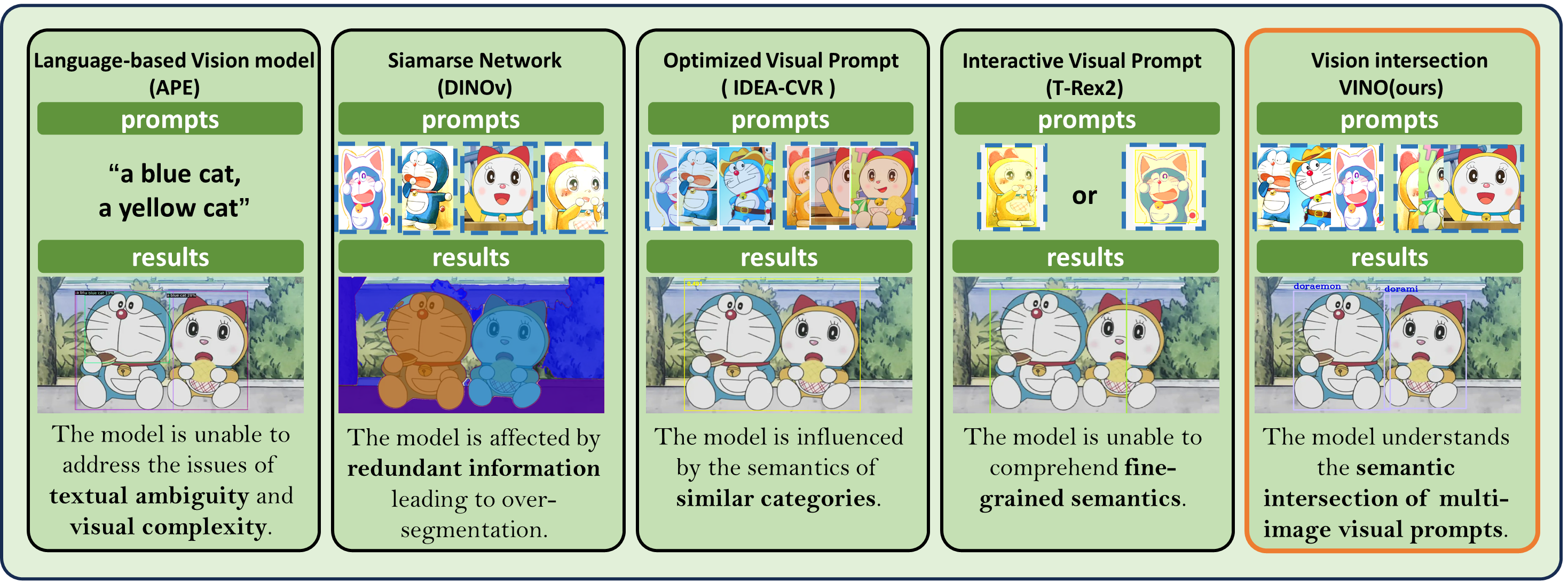} 
\caption{Comparison of various object detection models under visual and textual prompts. The figure highlights the challenges faced by existing models such as language-based vision models, siamese networks, optimized visual prompts, and interactive visual prompts, including issues with textual ambiguity, redundant information, semantic overlap, and fine-grained comprehension. In contrast, Vision Intersection Network (VINO) effectively addresses these challenges by leveraging the semantic intersection of multi-image visual prompts, enhancing detection accuracy and generalization in open set environments.}
\label{fig1}
\end{figure*}

\begin{abstract}
Open Set Object Detection has seen rapid development recently, but it continues to pose significant challenges. Language-based methods, grappling with the substantial modal disparity between textual and visual modalities, require extensive computational resources to bridge this gap. Although integrating visual prompts into these frameworks shows promise for enhancing performance, it always comes with constraints related to textual semantics. In contrast, viusal-only methods suffer from the low-quality fusion of multiple visual prompts. In response, we introduce a strong DETR-based model, Visual Intersection Network for Open Set Object Detection (VINO), which constructs a multi-image visual bank to preserve the semantic intersections of each category across all time steps. Our innovative multi-image visual updating mechanism learns to identify the semantic intersections from various visual prompts, enabling the flexible incorporation of new information and continuous optimization of feature representations. Our approach guarantees a more precise alignment between target category semantics and region semantics, while significantly reducing pre-training time and resource demands compared to language-based methods. Furthermore, the integration of a segmentation head illustrates the broad applicability of visual intersection in various visual tasks. VINO, which requires only 7 RTX4090 GPU days to complete one epoch on the Objects365v1 dataset, achieves competitive performance on par with vision-language models on benchmarks such as LVIS and ODinW35. 
\end{abstract}

%

\section{Introduction}

Open set Object Detection has gained significant attention in recent years due to the vast diversity of objects in the real world, which traditional closed set detection methods struggle to handle\cite{zhu2020deformable}. The main considerations of open set object detection include the capability to localize proposal regions effectively and the alignment of object semantics with region semantics. 

Previous Open Set Object Detection methods have predominantly relied on textual labels as semantic anchors, which often fail to capture the complete semantic information of diverse object categories. Some approaches\cite{liu2023grounding,shen2024aligning} have attempted to enrich object semantics through descriptive sentences, but face challenges as the number of categories increases. When dealing with numerous categories, textual descriptions often struggle to convey detailed visual information. Additionally, the longer queried texts associated with a wide range of categories enhance the difficulty for large language models of these approaches to effectively understand textual semantics. In contrast, by leveraging the descriptive power of visual prompts, MQ-Det\cite{xu2024multi} employs adapters to fine-tune text encoder, facilitating the fusion of visual information with textual semantics. Although MQ-Det benefits from the rich semantics of images via visual prompts, its reliance on textual semantics, which is constrained by the extensive parameter size of Large Language Models, limits its ability to fully exploit this visual information.

Visual-only methods\cite{kang2019few, han2022few} typically utilize a two-branch Siamese network architecture to align visual prompts with the target image. However, these approaches are limited by the generalization capabilities of the Siamese network and are primarily used in few-shot object detection. Inspired by in-context prompting in Large Language Models, DINOv\cite{li2024visual} employs visual in-context prompts to enhance semantic understanding of targets. However, DINOv retains only single-step semantic information, making it difficult to fully describe the comprehensive details of a category, and results in a performance decline.

A picture paints a thousand words. More pictures say more. To enhance the class-level semantic intersection learning capability of Open Set Object Detection, we developed a new region classifier architecture model, Visual Intersection Network for Open Set Object Detection (VINO). VINO retains semantic information across all time steps by utilizing a multi-image visual bank and a novel mechanism for updating multi-image prompts. This innovative approach allows VINO to process multiple visual inputs simultaneously, enhancing the model's capability to understand and represent category-specific features accurately and robustly. By employing multiple image intersections as semantic anchors, our model not only overcomes the limitations of textual and single-step image descriptions but also bridges the gap between cropped images and the full image context. The dynamic capability of the multi-image visual bank to continuously update and optimize allows VINO to integrate new information flexibly and refine feature representations effectively, ensuring strong generalization even when encountering unseen objects. A comparison between our work and previous work is shown in Fig \ref{fig1}.

By pre-training on the Objects365v1 dataset and evaluating on the ODinW-35 and LVIS datasets, VINO has achieved competitive performance compared to existing vision-language and vision-only methods. To verify the general applicability of semantic intersections in enhancing label semantics, we added a segmentation head to the model. By pre-training VINO on the COCO dataset, the segmentation results are comparable to current methods, demonstrating the broad applicability of semantic intersections in visual tasks. In summary, our contributions are as follows:
\begin{itemize}
\item We are the first to propose using multi-image semantic intersections in the field of Open set Object Detection, and our model achieve comparable performance with vision-language and vision-only methods.
\item  We design and build the Visual Intersection Network for Open Set Object Detection (VINO), constructing a multi-image visual bank and introducing a multi-image updating mechanism. By retaining  semantic information from all time steps and  learning the semantic intersections from the multi-image visual bank, VINO ensure the quality of multi-visual prompts.
\item We conduct extensive experiments and visualization analyses, demonstrating our model’s ability to handle open set object detection tasks. Specifically, VINO achieved an AP\(^b\) of 38.1 on Obj365 v1, 29.2 on the LVIS v1 validation set, and 24.5 on the ODinW-35 validation set, outperforming GLiP by 2.3 points on the LVIS v1 validation set and by 1.1 points on the ODinW-35 validation set.
\end{itemize}

\section{Related Work}
\subsection{Open-Vocabulary Object Detection}
With the emergence of large pre-trained vision-language models like CLIP\cite{radford2021learning} and ALIGN\cite{jia2021scaling}, methods based on vision and language\cite{kamath2021mdetr, gu2021open, zhang2022glipv2, zhang2023simple,yan2023universal} have gained significant popularity in the field of open-vocabulary object detection. These methods locate objects using language queries while effectively handling open set problems. OV-DETR\cite{zang2022open} is the first end-to-end Transformer-based open-vocabulary detector, combining CLIP embeddings from both images and text as object queries for the DETR decoder. GLIP\cite{li2022grounded} treats object detection as a grounding problem and achieves significant success by semantically aligning phrases with regions. To address the limitations of single-stage fusion in GLIP, Grounding DINO\cite{liu2023grounding} enhances feature fusion at three stages: neck, query initialization, and head phases, thus tackling the issue of incomplete multimodal information fusion. Furthermore, APE\cite{shen2024aligning} scales the model prompts to thousands of category vocabularies and region descriptions, significantly improving the model’s query efficiency for large-scale textual prompts. The language-based models aim to enhance the semantic description of language queries to adapt to various visual environments, achieving remarkable progress in zero-shot and few-shot settings. However, relying solely on text poses limitations due to language ambiguity and potential mismatches between textual descriptions and complex visual scenes. This underscores the ongoing need for improved integration of visual inputs to achieve more accurate and comprehensive results. Recent advancements suggest that incorporating richer visual prompts and enhancing multimodal fusion techniques are crucial for overcoming these challenges and pushing the boundaries of open-vocabulary object detection further.

\subsection{Object Detection by Visual Queries}
Expanding on language-based object detectors, some methods\cite{zhou2022conditional, zhou2022learning}have introduced visual elements to enhance detection accuracy and semantic richness. MQ-Det\cite{xu2024multi} utilizes image examples as visual prompts to enhance textual semantics, thereby enabling more effective open-vocabulary object detection. However, it remains constrained by textual semantics. Additionally, some methods\cite{kang2019few,han2022few} explore the possibility of object detection using only visual prompts.  It primarily address few-shot object detection\cite{fan2020few, han2021query, han2022meta} and typically employ a two-branch Siamese network. For example, FCT\cite{han2022few} uses a two-branch Siamese network to process target images and visual queries in parallel, computing the similarity between image regions and a few examples for few-shot object detection. OWL-ViT\cite{minderer2022simple} leverages CLIP's parallel paradigm and uses detection datasets for fine-tuning to adopt image examples for one-shot image-conditioned object detection. Similarly, DINOv\cite{li2024visual} expands on this concept by employing visual instructions (such as boxes, points, masks, doodles, and specified regions referencing another image) to handle open set segmentation. These visual methods often adopt a Siamese network architecture, which has limitations in zero-shot learning capability. To address these limitations and improve semantic understanding, our goal is to learn the semantic intersection of multiple images. VINO enriches visual semantics by retaining semantic information in all time steps using a multi-image visual bank. Our approach not only improves the model's ability to understand complex visual scenes but also enhances its robustness and generalization in open set scenarios.

\section{Method}
\begin{figure*}[t]
\centering
\includegraphics[width=0.8\textwidth]{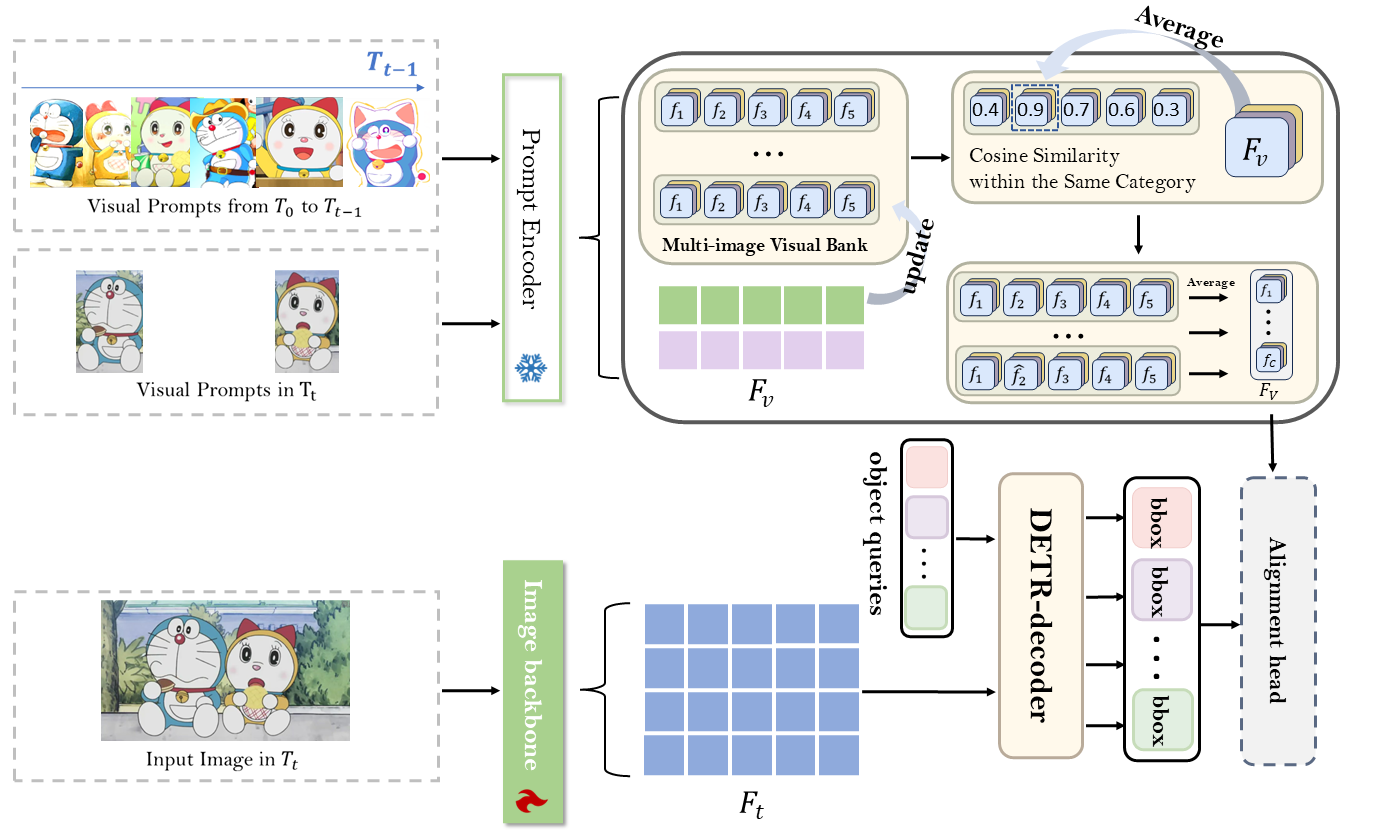} 
\caption{The model architecture of VINO with multi-image visual bank.The VINO model architecture incorporates a visual prompt encoder that extracts features from cropped images in \( T_{t-1} \) as visual prompts and stores them in the multi-image visual bank. When a new target image is processed, the model uses labels to get visual prompts and the prompt encoder to extract relevant features. Through cosine similarity-based selection and feature updating, the multi-image visual bank maintains and refines semantic intersections across categories, thereby improving the detection and alignment of objects in the target image.}
\label{fig2}
\end{figure*}

This section describes our proposed model VINO: a DETR-like detection framework that retains the semantic intersection of images in all time steps. This allows the model to learn the semantic intersections related to categories, enhancing the model's semantic alignment capability and detection performance. First, we introduce the key component of our model, the multi-image visual bank, which serves as the foundation for our approach. This is followed by a detailed overview of the overall framework of VINO.

\subsection{Multi-image Visual Bank}
\textbf{Rethinking Features in the Multi-image Visual Bank}: Our goal is to address the insufficiency of category semantics in a single time step by leveraging the semantic intersections across categories. These intersections capture common features and nuances, providing a more robust semantic representation. However, as the number of input images increases, retaining all the semantics of the label becomes impractical due to significant memory consumption. A straightforward FIFO (first-in, first-out) approach would result in the loss of valuable semantic information from previous time steps, which is unacceptable for maintaining accurate category descriptions over time. 

As images of the same category continuously appear, they create a stream of visual prompts. While this can be seen as a continuous influx of information, the challenge lies in retaining essential semantics within limited memory constraints. We address this by constructing a multi-image visual bank and employing multi-image updating mechanism. Our approach allows us to compress and retain critical semantic information across all time steps without overwhelming memory resources. The visual bank dynamically processes ROI features and maintains a prompt feature library. It selectively retains the most representative semantic information by averaging and aligning new features with existing ones, thus preserving the core semantics of each category. Our method ensures that, even as visual prompts accumulate, the memory bank remains efficient and semantic integrity is maintained.

\begin{algorithm}[tb]
\caption{Prompt Update Algorithm for Multi-image Visual Bank}
\label{alg:prompt_updating}
\textbf{Input}: ROI Features $F_{\text{prompt}}$: new feature to be integrated, ROI Category $I_{\text{prompt}}$: category of the new feature, Multi-image visual bank \( F_b = \{f_i||f_i|=n\}_{i=1}^{|C|}\)\\
\textbf{Parameter}: Defined maximum length $n$ for each semantic category in the multi-image visual bank\\
\textbf{Output}: Updated visual bank
\begin{algorithmic}[1] 
\FOR{each category in $F_b$}
\IF{category.label == $I_{\text{prompt}}$}
\STATE features = category.features
\FOR{$i = 1$ to $n$}
\IF{features[$i$] == 0}
\STATE features[$i$] = $F_{\text{prompt}}$
\STATE \textbf{return} $F$
\ENDIF
\ENDFOR
\FOR{$m=1$ to $n$}
\STATE $s_{m}$ = cosine\_similarity($F_{\text{prompt}}$, features[$m$])
\ENDFOR
\STATE $k$ = \texttt{argmax}($s_m$)
\STATE features[$k$] = average($F_{\text{prompt}}$, features[$k$])
\ENDIF
\ENDFOR
\STATE \textbf{return} $F$
\end{algorithmic}
\end{algorithm}

\noindent\textbf{Initialization and updating of the Multi-image Visual Bank}: During initialization, all entries in the multi-image visual bank are set to zero. Formally, the multi-image visual bank is represented as \( F_b = \{f_i||f_i|=n\}_{i=1}^{|C|}\), where \( f_i \in \mathbb{R}^{n \times d} \), \( |C| \) represents the number of categories, \( n \) is the number of visual prompts, and \( d \) is the dimension of the visual features. This initial state ensures a clean slate, ready to incorporate meaningful features as they are processed. When new features \( F_{v} \) are received, they are integrated into the corresponding \( f_{i} \) based on their category \( I_i \). The integration process (as Algorithm \ref{alg:prompt_updating}) is carefully designed to ensure efficient and effective updating of the visual bank while maintaining the semantic intersections of each category.

\noindent\textbf{Direct Replacement of Zero Entries}: If any sub-feature in \( f_{i} \) is zero, it indicates that this slot is currently unused. The new feature \( F_{v} \) is directly placed into this slot, ensuring all slots are utilized as new data arrives.

\noindent\textbf{Similarity-Based calculation}: If all sub-features in \( f_{i} \) are non-zero, a more efficient approach is required to integrate the new feature without losing valuable information from previous time steps. To achieve this, we calculate the cosine similarity between \( F_{v} \) and each sub-feature in \( f_{i} \). At the same time, each sub-element \( f_{[i,k]} \) represents the k-th view. The cosine similarity \( s_m \) for the  m-th sub-feature is computed as:

\begin{equation}
s_m = \cos(F_{v}, f_{[i, m]}) \quad m \in [1, n].
\end{equation}

This step identifies the sub-feature that is most similar to the new feature, indicating redundancy or relevance in the semantic space.

\noindent\textbf{Averaging and Updating}: Once the sub-feature with the highest cosine similarity is identified (denoted as \(k = gmax(s_m)\)), we update this sub-feature by averaging it with the new feature \( F_{\text{prompt}} \)
\begin{equation}
\hat{f}_{[i, k]} = \text{Average}(F_{v}, f_{[i, k]}).
\end{equation}
In this context, \( \hat{f}_{[i, k]} \) represents the new value of the sub-feature after incorporating the information from \( F_{\text{prompt}} \). This averaging process helps in retaining both the new and existing semantic information, thereby preserving temporal context and reducing noise. By adopting this method, our multi-image visual bank dynamically processes and retains essential semantic information across all time steps. This allows our model to effectively leverage semantic intersections, providing robust and comprehensive category representations even with limited memory resources. This design ensures that the bank remains efficient and scalable, accommodating new categories and seamlessly evolving visual data.

\subsection{The framework of VINO}
Our model consists of several key components: the \textbf{Image Backbone}, a visual encoder that extracts features from the target image; the \textbf{DETR Decoder}, a visual decoder that identifies the location and semantic information of proposed regions; the \textbf{Prompt Encoder}, which extracts features from visual prompts; and the \textbf{Multi-image Visual Bank}, a memory bank that preserves visual prompt information for each category and outputs their semantic intersections. By aligning the semantic information of the proposed regions with the semantic intersections of the visual prompts, we can assign labels to each proposed region. Our goal is to identify objects of interest in a target image \( I_t \).

Specifically, the model takes the target image \(I_t \in \mathbb{R}^{3 \times h \times w}\) and the set of labels \(R = \{r_1, r_2, \ldots, r_{|R|}\}\) as input. Here, \(r_i = (x_1, y_1, x_2, y_2, I_i) \in \mathbb{R}^{5}\) represents the coordinates of the top-left and bottom-right corners, along with the corresponding category label.

\noindent\textbf{Feature Extraction and Region Proposal}: For the target image \(I_t\), the initial step involves feature extraction using the Image Backbone to produce the feature representation \(F_t\):
\begin{equation}
F_t = \text{Image-Backbone}(I_t)
\end{equation}
where \(F_t \in \mathbb{R}^{bs \times D}\), with \(bs\) representing the batch size and \(D\) denoting the dimensionality of the feature vectors.

To enable the model to flexibly accommodate new and unseen categories, we employ a DETR-like decoder to process the extracted features \(F_t\). The decoder leverages object queries \(F_{q}\in \mathbb{R}^{bs \times n_q \times D}\) as prompts, where \(n_q\) represents the number of object queries used. These object queries serve as learnable parameters that guide the model in identifying potential object regions within the target image.
\begin{equation}
B, F_{r}=\text{DETR-Decoder}(F_t,F_q)
\end{equation}
The DETR-like decoder operates by decoding the features \(F_t\) into two outputs: the coordinates of the proposed regions \(B \in \mathbb{R}^{bs \times n_q \times 4}\) and the corresponding feature representations of these proposed regions \(F_{r} \in \mathbb{R}^{bs \times n_q \times D}\). 

To further validate the broad applicability of visual intersections in visual tasks, we extend the model by incorporating a segmentation head. This addition allows the model to also output predicted masks \(M \in \mathbb{R}^{bs \times n_q \times h \times w}\). By leveraging the semantic intersection mechanism, the model is capable of not only detecting objects but also segmenting them within complex visual scenes, thereby demonstrating the versatility of our approach.

\noindent\textbf{Feature Fusion}: For the set of labels \(R\), we first use the Prompt Encoder to extract the features from each region:
\begin{equation}
F_{v} = \text{Prompt-Encoder}(R, I_t)
\end{equation}
This step produces the feature representation \(F_{v}\), which captures the semantic information of the regions associated with the labels in the target image \(I_t\).

Next, we perform feature fusion by updating the multi-image visual bank \(\hat{F}_{i}\) with the features extracted from the regions, aligning them with the same category in the visual bank, as described in the previous section:

\begin{equation}
\hat{F}_{i} = \text{Prompt-Update}(F_{v}, F_{i})
\end{equation}

This fusion process integrates the new region features into the existing visual bank, ensuring that the updated bank retains and reflects the latest semantic information.

After the fusion, we apply a Multi-Layer Perceptron (MLP) to the averaged and dimension-aligned features to obtain the final average feature representation \(F_{V}\in R^{bs\times |C|\times D}\):
\begin{equation}
F_{V} = \text{MLP}(\text{Average}(\hat{F}_{i}))
\end{equation}
\noindent\textbf{Label Assignment}: Finally, we use the Alignment Head to match the features of the proposed regions \(F_{r}\) with the averaged features \(F_{V}\) to determine the semantic labels:

\begin{equation}
R = \text{Softmax}(F_{r} @ F^T_{V})
\end{equation}

This step outputs \(R \in \mathbb{R}^{bs \times n_q \times C}\), assigning the most probable semantic labels to each proposed region.

\section{Experiments}
\subsection{Setup}
\textbf{Dataset and Settings}. To rigorously evaluate the performance of our open set object detector, we introduce VINO-D, which leverages pre-training on the Objects365v1 dataset\cite{shao2019objects365}. This dataset encompasses a comprehensive collection of 600K images spanning 365 object categories, marked with 30 million bounding boxes. For open set detection evaluation, VINO-D is tested on two benchmarks. Firstly, we employ the LVIS v1 validation set\cite{gupta2019lvis}, known for its long-tail distribution of 1,203 object categories. Secondly, on the ODinW35 dataset\cite{li2022grounded}, which consists of 35 diverse datasets designed to challenge model performance in varied real-world scenarios. This dataset includes many rare categories seldom represented in training datasets, providing a stringent test of our model's transferability and effectiveness across common and rare object categories. In addition to VINO-D, we develop VINO-S by integrating a segmentation head to expand its capabilities to both detection and segmentation tasks. This model is pre-trained on the COCO2017 dataset\cite{lin2014microsoft}, which comprises approximately 110K images used for both object detection and panoramic segmentation. VINO-S is meticulously evaluated on the LVIS v1 validation set, demonstrating the broad application prospects of semantic intersections of visual prompts in visual tasks. \\
\textbf{Training Details}. Both VINO-D and VINO-S utilize APE-D weights for processing target images, with ViT-L\cite{fang2024evaMQ} serving as their mage backbone. The prompt encoder is CLIP-L, which remains frozen during training. The number of prompts is set to 5, and the number of object queries is set to 900. Training is conducted on 2 RTX4090 GPUs, with a batch size of 1, using the AdamW optimizer with a learning rate of 5e-5. VINO-D is pre-trained on the Objects365v1 dataset for 1 epoch, which takes approximately 7 RTX4090 GPU days. Similarly, VINO-S is pre-trained on the COCO2017 dataset for 1 epoch, requiring about 2 RTX4090 GPU days. To address the substantial domain shift caused by the prompt encoder taking cropped images as inputs\cite{li2024visual}, we control the resolution of these cropped images to ensure high-quality vision prompts. Specifically, the resolution of the first prompt image is maintained at no less than 2000, while subsequent visual prompts have resolutions no lower than 1600.
\subsection{Visual Intersection Open Detection and segmentation}

\begin{table*}[t]
\centering
\begin{tabular}{ c c c c c c c c}
\hline
\multirow{2}{*}{\textbf{Method}} & \multirow{2}{*}{\textbf{Backbone}} & \multirow{2}{*}{\textbf{Semantic Data}} & \multirow{2}{*}{\textbf{Type}} & \multicolumn{1}{c}{\textbf{objects365}} & \multicolumn{1}{c}{\textbf{LVIS-v1 val}} & \multicolumn{2}{c}{\textbf{Odinw-35 val}} \\ 
 &  &  &  & AP\(^b\) & AP\(^b\) & AP\(^b_\text{average}\) & AP\(^b_\text{median}\) \\ \hline
OWL             & ViT-L   & O365+VG+...        & Text Open set   & -      & 34.6  & 18.8  & 9.8   \\
GLIP            & Swin-L  & FourODs+... & Text Open set   & 36.2   & 26.9  & 23.4  & 11    \\ 
UNINEXT         & ViT-H   & O365v2+COCO+... & Text Open set   & 23     & 14    & -     & -     \\ 
OpenSeeD-L      & Swin-L  & O365v2+COCO+... & Text Open set   & -      & 23    & 15.2  & 5     \\ 
DINOv (L)       & Swin-L  & SAM+COCO+...      & Visual Prompt   & -      & -     & 15.7  & 4.8   \\ 
MQ-GLIP-L       & Swin-L  & O365           & Text and visual & -      & 34.7  & 23.9 & -     \\ 
VINO-D(ours) & ViT-L & O365   & Visual Prompt   & 38.1 & 29.2 & 24.5  & 9.4 \\ \hline
\end{tabular}
\caption{Open set segmentation results for different methods. “–” indicates that the work does not have a reported number.}
\label{table1}
\end{table*}
\begin{table*}[t]
\centering
\begin{tabular}{c c c c c c c c}
\hline
\multirow{2}{*}{\textbf{Method}} & \multirow{2}{*}{\textbf{Backbone}} & \multirow{2}{*}{\textbf{Semantic Data}} & \multirow{2}{*}{\textbf{Type}} & \multicolumn{2}{c}{\textbf{COCO}} & \multicolumn{2}{c}{\textbf{LVIS v1 val}} \\ 
 &  &  &  & AP\(^b\) & AP\(^m\) & AP\(^b\) & AP\(^m\) \\ \hline
GLIPv2        & Swin-H   & O365+COCO+... & Text Open set  & 64.1  & 47.4  & -     & -      \\ 
UNINEXT       & ViT-H    & O365v2+COCO  & Text Open set  & 60.6  & 51.8  & 14    & 12.2   \\ 
APE (D)       & ViT-L    & O365v2+COCO+... & Text Open set  & 58.3  & 49.3  & 59.6  & 53     \\
DINOv (L)     & Swin-L   & COCO  & Visual Prompt  & 54.2  & 50.4  & -     & -      \\ 
VIOSD-S (ours) & ViT-L    & COCO  & Visual Prompt  & 60.9  & 48.1  & 13.4  & 13.6 \\ \hline
\end{tabular}
\caption{Open set segmentation results for different methods. “–” indicates that the work does not have a reported number.}
\label{table2}
\end{table*}

\subsubsection{Object Detection}
In \textbf{Table \ref{table1}}, we present the detection results of our VINO-D model, which is pre-trained on Objects365v1 and evaluated on well-established benchmarks, including LVIS v1 val and ODinW35. VINO-D achieves the best or comparable results across these datasets. Specifically, VINO-D reaches an AP\(^b\) of 38.1 on Objects365v1, 29.2 on LVIS v1 val, and 24.5 on ODinW35.

Compared with current vision-language methods (GLIP, UNINEXT), VINO-D achieves competitive results. For instance, while GLIP achieves high AP\(^b\) on Objects365 using language queries, VINO-D performs exceptionally well with vision queries, showcasing the robustness of semantic intersections learned from multiple images. This ability of semantic intersection enables our model to have high detection accuracy without relying on text information. Furthermore, VINO-D’s performance on the LVIS v1 validation set is noteworthy. It achieves an AP of 29.2, which is on par with or better than several state-of-the-art vision-language methods. This result underscores the efficiency of using visual prompts to enhance semantic understanding and object detection performance even in a large number of categories. 

Compared to current visual methods, VINO-D significantly outperforms DINOv(L) by 8.8 points and MQ-GLIP-L by 0.6 points in terms of AP\(^b\) on the ODinW35 dataset. This showcases the effectiveness of our approach in handling domain shifts through the semantic intersections via a multi-image visual bank. DINOv(L) illustrates the challenge of domain shifts specifically caused by the differences in resolution size between cropped and target images. Meanwhile, MQ-GLIP-L, despite employing visual prompts to enhance textual semantics, is limited by the inherent constraints of text-based representations. Together, these observations highlight two challenges in domain adaptation: resolution discrepancies and text-based limitations, which our method successfully addresses. In particular, MQ-GLIP-L surpasses our method by 5.5 points on the LVIS dataset; however, it underperforms in real-world scenarios, as evidenced by its lower performance compared to our results. This suggests that MQ-GLIP-L's reliance on textual semantics restricts its generalization capabilities in diverse environments. Our approach, on the other hand, leverages the semantic intersection of multiple visual prompts, allowing VINO-D to maintain high performance across various and challenging detection scenarios. 

\subsubsection{Object Segmentation}
In \textbf{Table \ref{table2}}, we present the segmentation results of our VINO-S model, which includes a segmentation head. This model is pre-trained for detection and panoramic segmentation on the COCO dataset and evaluated for segmentation performance on the LVIS v1 validation set. VINO-S achieves an \(AP^b\) of 60.9 on COCO, surpassing UNINEXT by 0.3 points in detection and GLIPv2 by 0.7 points in segmentation, thereby achieving comparable results to current mainstream vision-language and vision-only methods. On the LVIS v1 validation set, VINO-S achieves a segmentation \(AP^b\) of 13.4 and an \(AP^m\) of 13.6. This performance is particularly noteworthy as VINO-S outperforms UNINEXT. The 1.4-point improvement in segmentation \(AP^b\) highlights the effectiveness of our approach.

These results confirm that the semantic intersection facilitated by our multi-image visual bank significantly boosts performance in segmentation tasks. The ability of VINO-S to achieve high accuracy in both detection and segmentation tasks highlights the robustness and versatility of our approach. It successfully generalizes across different datasets and tasks, illustrating the broad applicability and effectiveness of semantic intersections in addressing diverse visual challenges in open set scenarios. This method  which leverages the semantic intersection of visual prompts to improve class-level semantic, serves as a powerful tool for tackling a wide array of visual tasks.
\subsection{Ablations Experiments}
\begin{table}[h]
\centering
\begin{tabular}{c c}
\hline
\textbf{Prompt Num} & \textbf{AP\(^b\)} \\ \hline
1  & 20.9 \\ 
5  & 29.2 \\ 
10 & 30.3 \\ 
20 & 31.3 \\ \hline
\end{tabular}
\caption{LVIS v1 val AP\(^b\) Scores for different prompt numbers}
\label{table3}
\end{table}

\noindent\textbf{Effectiveness of the Number of Vision Prompts}: We conduct an ablation study to evaluate the effectiveness of different numbers of vision prompts on the performance of VINO-D. The model is pre-trained on Objects365v1 using a fixed number of 5 vision prompts and then evaluated on the LVIS v1 validation set with varying numbers of prompts: 1, 5, 10, and 20. The results, presented in the table \ref{table3}, demonstrate several key insights: 

1. \textbf{Incremental Improvement and Diminishing Returns}: As the number of prompts increases from 1 to 20, there is a clear improvement in detection AP, from 20.9 with a single prompt to 31.3 with 20 prompts. However, the rate of improvement decreases as the number of prompts increases. 

2. \textbf{Significance of Multiple Prompts}: The substantial improvement from 1 to 5 prompts underscores the importance of using multiple prompts. It demonstrates that leveraging multiple vision prompts enables the model to better generalize and understand the various visual features within each category.

\begin{table}[h]
\centering
\begin{tabular}{ c c}
\hline
\textbf{Method} & \textbf{AP\(^b\)} \\ \hline
FIFN    & 55.4 \\ 
Average & 60.9 \\ \hline
\end{tabular}
\caption{COCO AP\(^b\) Scores for different updating mechanisms}
\label{table4}
\end{table}

\noindent\textbf{Effectiveness of the updating Mechanism}:To evaluate the effectiveness of the updating mechanisms for the multi-image visual bank, we employ two different approaches while training VINO-S on the COCO dataset: averaging based on cosine similarity and first-in, first-out (FIFO). With the number of visual prompts set to 5, we observed the following results in terms of detection AP\(^b\) on COCO (as shown in Table 4): using cosine similarity averaging to update the visual bank significantly outperforms the FIFO method by 5.5 points. By retaining the temporal steps and selectively integrating new features based on their similarity to existing features, the cosine similarity averaging method preserves the semantic information across all time steps more effectively. This leads to a higher-quality semantic intersection, which enhances the model’s ability to generalize and maintain high detection performance.
\subsection{Visualization}
\begin{figure}[h]
\centering
\includegraphics[width=0.5\textwidth]{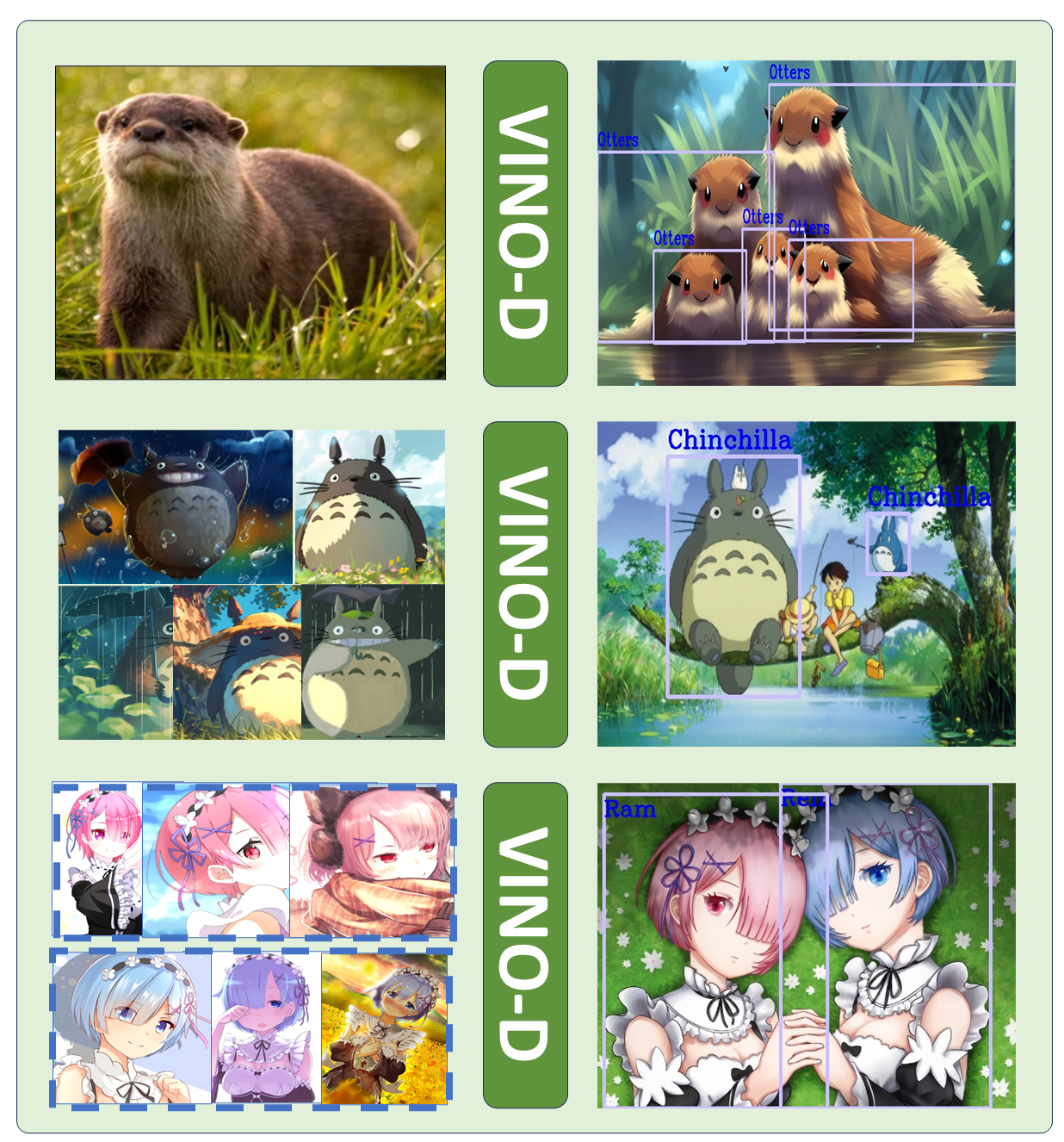} 
\caption{The Visualization of VINO-D.}
\label{fig3}
\end{figure}

The visualizations in Figure \ref{fig3} illustrate the superior capabilities of our model across a range of challenging scenarios. The first image demonstrates the model's proficiency in single-prompt detection, where a single visual prompt is sufficient to accurately identify and detect all instances of the target objects within the image. In the second image, the model effectively handles the detection of multiple instances across several categories within a single target image. This scenario highlights the robustness of the model’s semantic intersection approach, which allows it to maintain high detection performance even in complex environments with diverse object categories. The third example provides a clear illustration of the model's ability to distinguish between categories with similar semantic features. These examples collectively highlight the versatility and accuracy of our model in open set object detection tasks.

\section{Conclusion}

Through the innovative application of a multi-image visual bank, the VINO model demonstrates how mastering the semantic intersection of multi-image prompts can significantly boost the semantic understanding of object categories, thereby enhancing performance in open set object detection. By dynamically integrating and updating multiple visual prompts, VINO not only addresses the limitations associated with textual and single-image descriptions but also effectively narrows the contextual gap between cropped and full images. This ongoing refinement of feature representations ensures that VINO adapts flexibly to new information, achieving robust generalization capabilities even with unseen objects.  Additionally, we add a segmentation head to the model, demonstrating the generality of semantic integration in the visual domain. Experimental results show that VINO exhibits strong performance in open set object detection, achieving results comparable to current vision-language and vision-only methods. We hope that more studies will explore the application of semantic intersections in visual tasks, further expanding the capabilities and understanding of visual models in diverse and complex environments.
\bibliography{aaai25}
\section{Appendix}
\subsection{The Visualization of VINO-S}
\begin{figure}[h]
\centering
\includegraphics[width=0.5\textwidth]{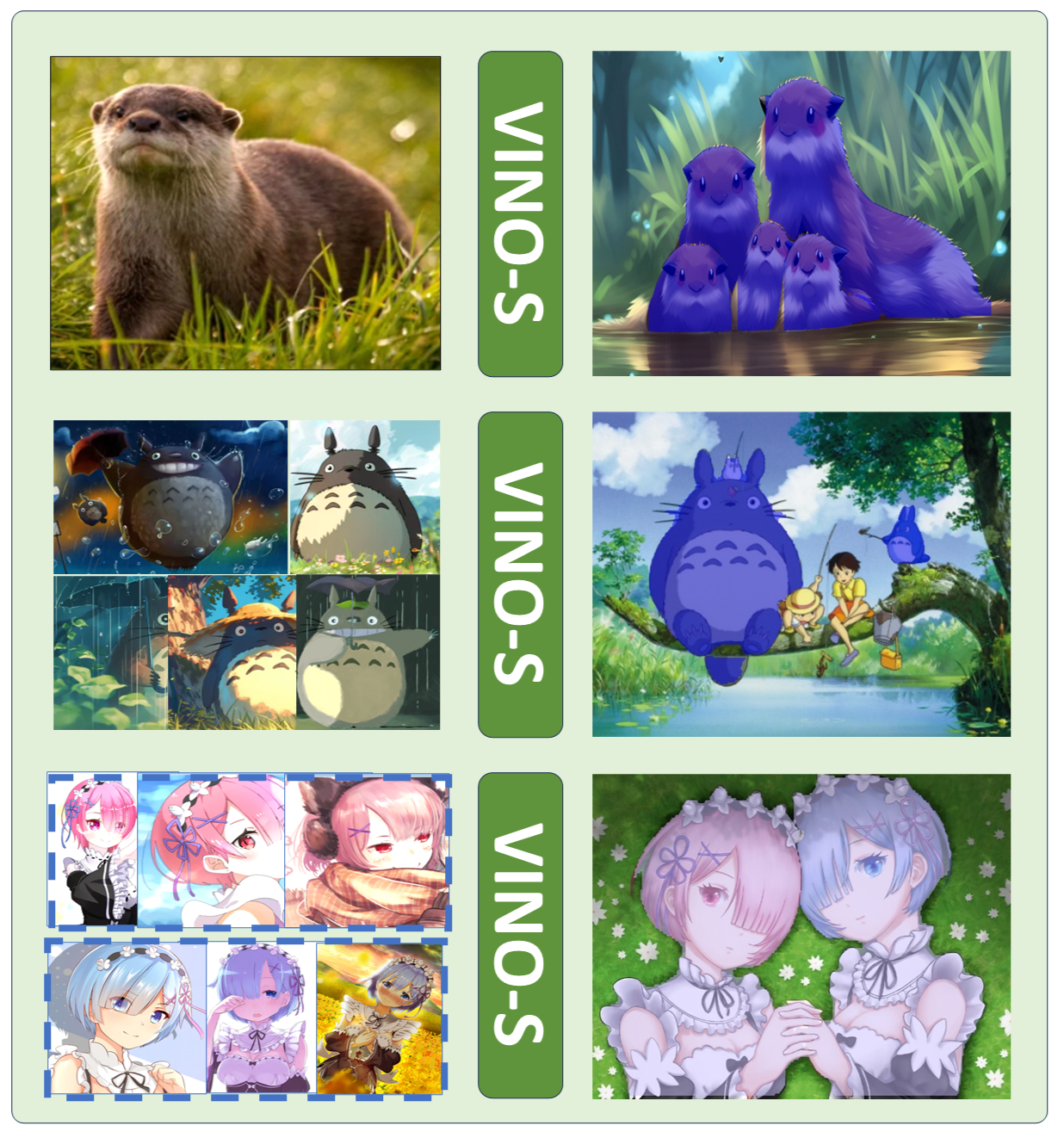} 
\caption{The Visualization of VINO-S.}
\label{fig1}
\end{figure}
The visualization of VINO-S, as shown in Figure \ref{fig1}, demonstrates that using the semantic intersection of visual prompts enables more effective segmentation of target semantics.
\subsection{Object Detection in the Wild}
Table \ref{tab:ODinW35} presents the zero-shot performance of VINO across 35 diverse datasets within the ODinW35 benchmark. These results demonstrate the model's robustness and adaptability in handling a wide range of real-world scenarios by semantic intersection. VINO consistently performs well across varying contexts and object categories, indicating its strong generalization ability in open-set object detection tasks.

\begin{table}[h]
    \centering
    \begin{tabular}{l r}
    \hline
    Dataset & AP\(^b\) \\ \hline
    AerialMaritimeDrone\_large & 9.383 \\
    AerialMaritimeDrone\_tiled & 22.368 \\ 
    AmericanSignLanguageLetters & 1.029 \\ 
    Aquarium\_Aquarium\_Combined & 30.308 \\ 
    BCCD\_BCCD & 20.261 \\ 
    boggleBoards\_416x416AutoOrient & 0.68 \\ 
    brackishUnderwater & 7.1 \\ 
    ChessPieces\_Chess\_Pieces & 15.291 \\ 
    CottontailRabbits & 80.9 \\ 
    dice\_mediumColor\_export & 2.418 \\ 
    DroneControl\_Drone\_Control & 17.307 \\ 
    EgoHands\_generic & 23.9 \\
    EgoHands\_specific & 7.39 \\ 
    HardHatWorkers\_raw & 2.781 \\ 
    MaskWearing\_raw & 1.247 \\ 
    MountainDewCommercial & 46.954 \\ 
    NorthAmericaMushrooms & 87.269 \\ 
    openPoetryVision & 0.00439 \\ 
    OxfordPets\_by-breed & 0.1434 \\ 
    OxfordPets\_by-species & 3 \\ 
    Packages\_Raw & 81.584 \\ 
    PascalVOC & 65.12 \\ 
    pistols\_export & 77.1 \\ 
    PKLot\_640 & 5.329 \\ 
    plantdoc\_416x416 & 1.17 \\ 
    pothole & 3.872 \\ 
    Raccoon\_Raccoon & 59.19 \\
    selfdrivingCar\_fixedLarge & 9.135 \\ 
    ShellfishOpenImages & 50.136 \\ 
    ThermalCheetah & 14.861 \\ 
    thermalDogsAndPeople & 42.696 \\ 
    UnoCards\_raw & 0.4911 \\ 
    VehiclesOpenImages & 65.452 \\ 
    websiteScreenshots & 0.3807 \\ 
    WildfireSmoke & 0.1283 \\ \hline
    \end{tabular}
    \caption{Zero-shot performance on ODinW-35}
    \label{tab:ODinW35}
\end{table}
\end{document}